\documentclass[letterpaper, 10 pt, conference]{ieeeconf}  

\IEEEoverridecommandlockouts                              

\usepackage{graphics} 
\usepackage{epsfig} 
\usepackage{amsmath} 
\usepackage{amssymb}  
\usepackage{array}
\usepackage{multirow}
\usepackage[font=small]{caption}

\usepackage[dvipsnames, table]{xcolor}

\title{\LARGE \bf
On the Feasibility of EEG-based Motor Intention \\ Detection for Real-Time Robot Assistive Control
}
\author{Ho Jin Choi$^{1\dagger*} $, Satyajeet Das$^{1\dagger}$, Shaoting Peng$^{1\dagger}$, Ruzena Bajcsy$^{1}$ and Nadia Figueroa$^{1}$
\thanks{$\dagger$ Equal contribution, listed in alphabetical order.}%
\thanks{$^{*}$Corresponding author: {\tt cr139139@seas.upenn.edu}}
\thanks{$^{1}$All authors are with School of Engineering and Applied Science, University of Pennsylvania, Pennsylvania, PA 19104 USA.}%
}

\begin{document}
\maketitle
\thispagestyle{empty}
\pagestyle{empty}
\begin{abstract}
This paper explores the feasibility of employing EEG-based intention detection for real-time robot assistive control. We focus on predicting and distinguishing motor intentions of left/right arm movements by presenting: i) an offline data collection and training pipeline, used to train a classifier for left/right motion intention prediction, and ii) an online real-time prediction pipeline leveraging the trained classifier and integrated with an assistive robot. Central to our approach is a rich feature representation composed of the tangent space projection of time-windowed sample covariance matrices from EEG filtered signals and derivatives; allowing for a simple SVM classifier to achieve unprecedented accuracy and real-time performance. In pre-recorded real-time settings (160 Hz), a peak accuracy of 86.88\% is achieved, surpassing prior works. In robot-in-the-loop settings, our system successfully detects intended motion solely from EEG data with 70\% accuracy, triggering a robot to execute an assistive task. We provide a comprehensive evaluation of the proposed classifier.
\end{abstract}
\vspace{-4pt}
\section{Introduction}
As robots evolve from automated tools to intelligent collaborators, the importance of comprehending and predicting human intentions has taken center stage in robotics research. Particularly in scenarios where robots offer essential physical aid to the elderly or individuals with disabilities, the prediction of intentions becomes pivotal in establishing responsive and harmonious interactions that seamlessly blend human needs with robotic assistance without discomfort \cite{Sciutti2018HumanizingHI}. 

The term ``human intention prediction" encompasses a robot's capability to deduce the actions or choices that a human is likely to undertake. Such predictive capability can allow robots to dynamically align their behaviors, responses, and maneuvers in a proactive manner. In this work, our focus is deliberately narrowed down to motor intention prediction, as the overarching term ``intention" encompasses a broad spectrum of meanings \cite{Kuli2003EstimatingIF}. Motor intention prediction refers to forecasting a human's velocity, location, or force trajectory within a confined time frame, employing the principles of state estimation. Moreover, this extends to foreseeing the states of objects subsequent to a brief period of human movement. The crux of motor intention prediction lies in the search for optimal solutions among multiple potential trajectories. Furthermore, if the individuals have mobility issues, then solely using movement kinematics to predict motor intention is insufficient. Interestingly, most individuals with limited mobility still preserve the ability to produce motor function-related neural activity comparable to those observed in healthy individuals. This occurs as the brain areas and the peripheral nervous system governing movement largely continue to operate. This insight has spawned an accelerated development of biosignal-based assistive human-machine interfaces \cite{ATHAVALE201722, Esposito2021-to,TANG2023108712}, where sensors that decode physiological signals are used to decipher the \textit{human's intention} and used as \textit{control inputs} to an assistive device, including exoskeletons, prostheses, rehabilitation and assistive robotic systems; enabling humans to control such assistive devices with their thoughts and movement intentions \cite{mohebbi_human-robot_2020}.

To this end, researchers have vastly explored the use of electroencephalography (EEG) \cite{Lakany2007UnderstandingIO}, which captures electrical signals emitted by the brain cortex's pyramidal cells, acting as dipoles \cite{kirschstein2009source}. While non-invasive EEG primarily captures surface information from the frontal lobe and motor cortex, key areas for signal transmission and reception, it may not directly access signals from deep brain structures like the thalamus. The absence of thalamic signals, however, does not significantly impede brain-computer interface (BCI) and brain-machine interfaces (BMI) functionality, as EEG effectively captures the relevant cortical activity necessary for control interfaces. In fact, the viability of non-invasive EEG for BCI/BMI applications has been repeatedly demonstrated, given its accessibility and minimal invasiveness \cite{Vrbu2022PastPA}.

The main challenge of decoding EEG signals from non-invasive devices is the signal to noise ratio, requiring a classical multi-stage processing pipeline such as: i) signal acquisition, ii) preprocessing, iii) feature extraction, iv) classification, and v) control interface \cite{stages}. However, recent results have shown impressive setups where individuals can control robotic arms to do grasping \cite{meng_noninvasive_2016} and household tasks solely from EEG activity \cite{lee2023noir}.  Such methods powered by Deep Learning (DL) techniques, however, \textbf{do not run in real-time}. Since EEG signals are noisy and non-stationary DL techniques require either vast amounts of data to train models specifically tailored for each human or take minutes to classify the intention of the human \cite{Dillen2022-ti}, creating unnecessary processing bottlenecks for the use of EEG signals in fluid and real-time human-robot assistive control.

\begin{figure*}[!tbp]
      \centering
      \includegraphics[width=0.875\textwidth]{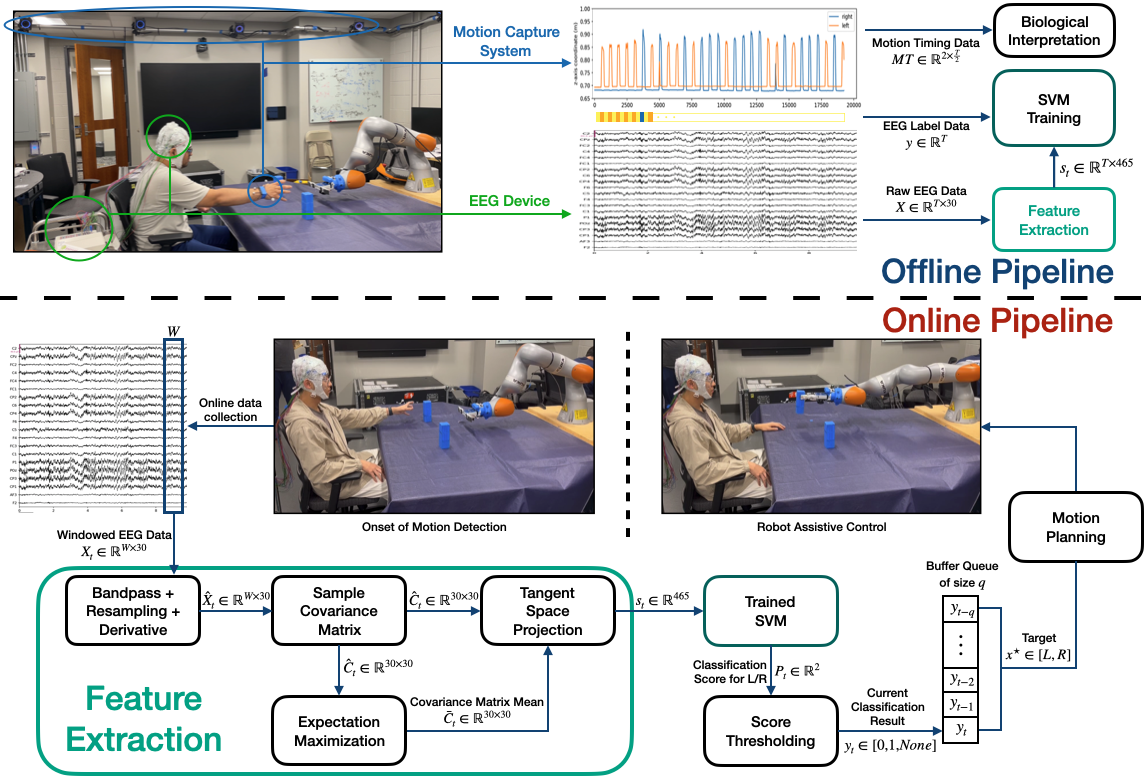}
      \caption{\textbf{Real-time EEG Motor Intention Prediction Framework}. In the offline pipeline, we first gather 4 minutes of raw EEG data as well as arm kinematics given by the motion capture system used to define the corresponding classification labels (left vs. right arm motion). In both offline and online pipelines we use the same feature extraction approach described in Section \ref{sec:methods} to process the EEG signals and construct a feature vector as the tangent space projection of the sample covariance matrices. An SVM classifier is trained per subject on 50\% of the collected data. The trained SVM classifier is then used online to predict left vs. right arm motion intention \textbf{solely from EEG data}, due to change of setting with the robot-in-the-loop a score thresholding phase is added to the pipeline to filter out noisy prediction and trigger the robot to provide assistance in reaching for an object with the left vs. right arm with high confidence.
      \label{fig:pipeline}}
      \vspace{-15pt}
  \end{figure*}

In contrast to such DL-based recent works, in this paper, we seek to develop an EEG signal decoding framework capable of predicting motion intention with \textbf{high accuracy, low data requirements and real-time performance}. To achieve this, we adopt a Riemannian geometry based classification approach in which segments of time-windowed EEG signals are represented as sample covariance matrices. Classification is then performed on the Riemannian manifold of symmetric positive definite matrices (SPD). By utilizing Riemannian operations to ensure that classification methods are faithful to the geometry of SPD manifolds when using covariance matrices as features, a myriad of recent works have shown promising results on classical BCI applications such as motor imagery, intention classification and P300 task \cite{100,7740054,ying_riemannian_2022,Congedo2017RiemannianGF,Rodrigues2019RiemannianPA,Barachant2012MulticlassBI,Barachant2013ClassificationOC},  Nevertheless, seldom works have demonstrated real-time classification performance in robot-in-the-loop applications from single subject training data. 

\textbf{Contributions:} In this paper, we introduce a real-time processing framework, that adopts the Riemannian geometry based EEG signal processing approach to extract rich covariance matrix features and predict motion intention (left vs. right arm movement) with a simple SVM classifier trained on 4 minutes of single-subject training data, as detailed in Section \ref{sec:methods} and depicted in Fig. \ref{fig:pipeline}. In pre-recorded offline real-time settings, i.e., predicting intention on collected testing data streamed in real-time referred to as the offline pipeline, we achieve a peak accuracy of 86.66\% at 160 Hz prediction frequency -- a considerable improvement to state-of-the-art performance in similar settings \cite{100}. We explore the feasibility of our trained classifier in an online robot-in-the-loop application referred to as the online pipeline, achieving 70\% accuracy with 1-2 seconds of queue as shown in Section \ref{sec:robot_exp_biological}. A comprehensive analysis of hyper-parameter and classification model selection is provided in Section \ref{sec:experiments}.

\section{Related works}
\label{sec:related_works}
\subsection{EEG features for classifiers}
One common approach to extracting features from EEG signals for classification is to solve the inverse problem of 3D source localization \cite{Michel2019EEGSI}. This method provides information about the brain region and its activation degree but typically requires a detailed model of the human head obtained from MRI scans or a canonical model. However, due to the computational complexity involved, this approach is often impractical for real-time or online applications.

As an alternative, researchers have sought to identify certain brain patterns now common in literature, particularly the  Readiness Potential (RP) and Event-Related Desynchronization/Synchronization (ERD/ERS) \cite{RP/ERD}. These patterns can be identified directly from raw EEG signals or their frequency representations obtained through techniques like the fast Fourier transform (FFT) or wavelet transform (WT). Moreover, the use of time derivatives alongside raw EEG signals has been investigated to capture temporal dynamics and enhance classification performance \cite{Andreou2016ComparingEI, Menceloglu2020SpectralpowerAR}. In addition to temporal features, the spatial relationships between EEG channels have been exploited to improve classification accuracy. Techniques such as spatial filters \cite{Menceloglu2020SpectralpowerAR}, graph structures \cite{Chen2022MentalSC}, and sample covariance matrices \cite{Barachant2013ClassificationOC} have been employed to capture the interplay between electrodes. These methods aim to leverage the spatial coherence of EEG signals across multiple channels, thereby providing valuable information for discriminating between different brain states.

By combining both temporal and spatial features, classifiers can effectively discern patterns in EEG signals associated with various cognitive states or tasks, facilitating applications in brain-computer interfaces, cognitive neuroscience, and clinical diagnostics.

\subsection{Online motor intention classification using EEG}
Traditionally, for human motion tracking and prediction, researchers have employed EMG-based techniques \cite{BI2019113, trigili2019detection}, camera-marker-based methods, or a fusion of the two. These methods excel in detecting precise positional and force-related information for different body segments but do not directly tap into the brain's motor intention. EEG presents a promising avenue for accessing rudimentary insights into movement through cortical potentials, including the readiness potential (RP) \cite{SCHURGER2021558, TRAVERS2020116286}. Debates exist around whether the onset of cortical potentials invariably indicates imminent physical action. However, it is generally observed that potentials occur approximately 500 ms before a voluntary motion starts, with a potential occurrence occurring 200 ms before such a motion is deemed inevitable \cite{Schurger2021WhatIT}. While efforts to detect RP for robotic control have shown promise, challenges persist, including the need for aggregating EEG signals across multiple trials and addressing noise and artifacts inherent in EEG data.

The advent of deep learning techniques has significantly advanced brain signal classification, with convolutional neural networks (CNNs) proving effective in deciphering nonstationary and nonlinear features in EEG data \cite{114}. Additionally, the fusion of neuromorphic computing and brain-computer interfaces (BCIs), particularly employing spiking neural networks (SNNs), has shown potential \cite{115}. However, these approaches often operate offline, limiting their applicability in real-time control scenarios.

For online BCI applications, simple classifiers like linear discriminant analysis (LDA) \cite{101}, Support Vector Machines (SVMs) \cite{102,103}, fuzzy logic-based classifiers \cite{105}, Gaussian \cite{106} and Bayesian \cite{107} classifiers that can deal with uncertainty, and shallow neural networks \cite{109,110,111,112} are more suitable. While more sophisticated fusion techniques combining SVMs, Random Forests, and Artificial Neural Networks have been proposed \cite{113}, they still primarily function offline.

Racz et al. \cite{100} pioneered online cortical potential classification on prerecorded data, attaining a 62.6\% binary classification accuracy by leveraging sample covariance matrix features and their underlying Riemannian geometry. However, the accuracy of online cases decreases compared to offline cases due to the inherent difficulty of single-trial detection and challenges in real-time artifact removal. Our proposed method now establishes the state-of-the-art standard, achieving an impressive 86.88\% in prerecorded real-time settings, and 70\% accuracy in real robot experiments.

\section{Methods}
\label{sec:methods}
We begin by introducing key preliminaries for our real-time EEG based intention prediction framework (Fig. \ref{fig:pipeline}): the sample covariance matrix feature and the SVM classifier. Next, we introduce the feature extraction part, followed by offline training and online execution pipelines.

\subsection{Preliminaries}
\label{sec:preliminaries}
\subsubsection{Sample Covariance Matrix}
In biosignal classification, we often deal with a multivariate signal $X\in\mathbb{R}^{W\times n}$, where $n$ represents the number of EEG channels and $W$ denotes window size. Understanding the spatial relationships between these channels is crucial, and this is where the sample covariance matrix, denoted as $C = Cov(X) =\frac{X^{T} X}{(n-1)}$, comes into play. When $W$ is considerably larger than $n$, this covariance matrix becomes a symmetric positive definite (SPD) matrix, placing it on the Riemannian manifold \cite{Barachant2012MulticlassBI} that captures the intrinsic geometric structure of SPD matrices. However, in cases where the sample covariance matrix is symmetric positive semi-definite, a small regularization term $\epsilon I $ is added, with $\epsilon$ being a very small positive constant, to the diagonal elements of the matrix. This ensures that the matrix becomes full rank and facilitates further analysis. 

Utilizing the Riemannian manifold has proven beneficial in biosignal classification tasks. Many works \cite{Barachant2013ClassificationOC, Barachant2012MulticlassBI, Congedo2017RiemannianGF, Nguyen2018EEGFD, Kalaganis2020ARG, Rodrigues2019RiemannianPA} have leveraged the Riemannian manifold to compute the mean of data points or covariance matrices, project all data points to the tangent space of the mean, and utilize the upper triangle of the tangent space points with traditional classifiers. By doing so, we preserve the geometric relationships between SPD matrices, reduce dimensionality, and maintain high classification accuracy.

\begin{figure}[!tbp]
  \centering
  \includegraphics[width=0.85\linewidth]{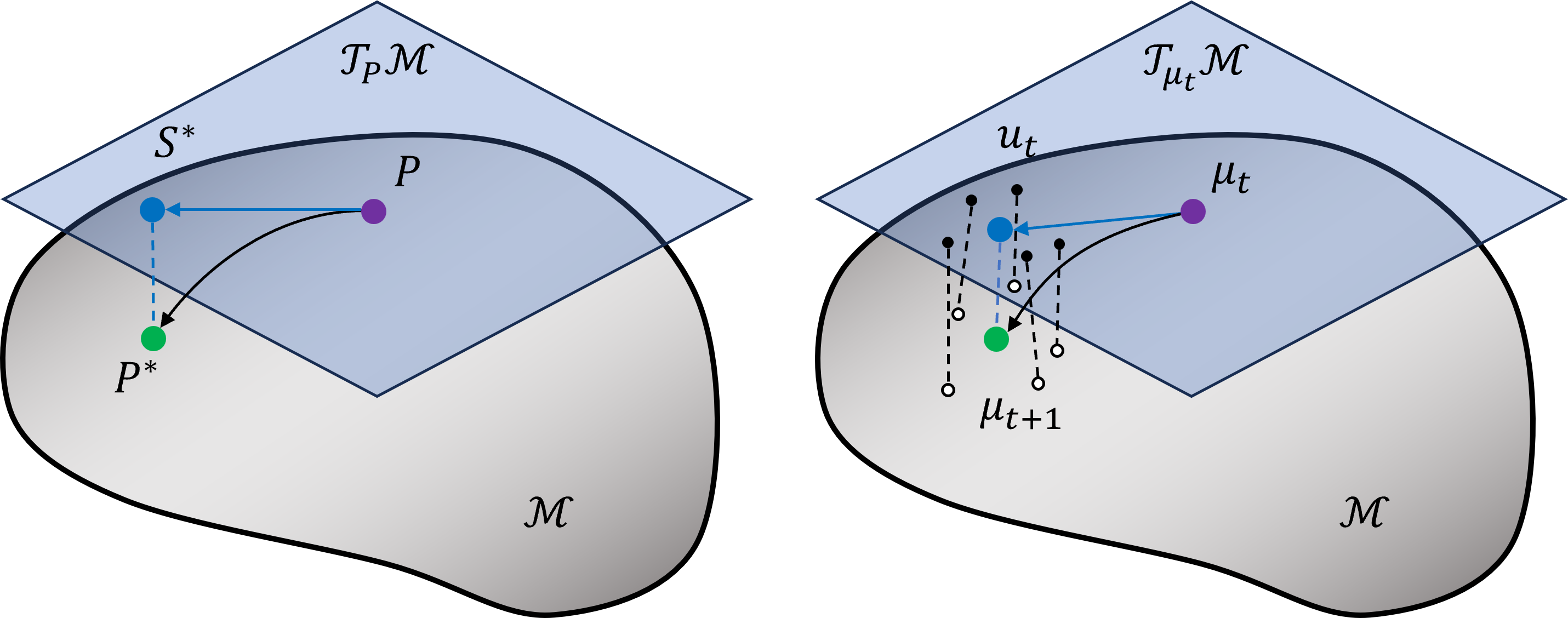}
  \caption{Illustration of a Riemannian manifold and the tangent space of a point. (left) Tangent space projection from $P^*$ to $S^*$ using $P$'s tangent space. (right) A step of the iterative EM process used to determine the mean of points on the Riemannian manifold.}
  \label{fig:tanget_space}
  \vspace{-15pt}
\end{figure}

The mean of N points on the manifold of the SPD matrix, denoted as $\mu$, serves as a reference point for tangent space projection. It is iteratively computed using the expectation-maximization method \cite{Calinon2019GaussiansOR} as depicted in Fig. \ref{fig:tanget_space}:
\begin{equation}
    u=\frac{1}{N}\sum_{i=1}^N \text{Log}_\mu (x_i), \quad \mu=\text{Exp}_\mu(u)
\end{equation}
Here, $\text{Log}_P$ denotes the Riemannian logarithm mapping for SPD matrix $P$, projecting a point to the tangent space of $P$, while $\text{Exp}_P$ represents the Riemannian exponential mapping for $P$, projecting a point back from the tangent space to the manifold. These mappings are defined as
\begin{equation}
\begin{split}
    & \text{Log}_P(P^*)=S^*=P^\frac{1}{2}log(P^{-\frac{1}{2}} P^* P^{-\frac{1}{2}})P^\frac{1}{2} \\
    & \text{Exp}_P(S^*)=P^*=P^\frac{1}{2}exp(P^{-\frac{1}{2}} S^* P^{-\frac{1}{2}})P^\frac{1}{2} \\
\end{split}
\end{equation}
where $P^*$ also lies on the SPD matrix manifold, and $S^*$  represents a point on the tangent space, which is a symmetric matrix, as illustrated in Fig. \ref{fig:tanget_space}. Then, the upper triangle of the tangent space points, denoted as $s^*=vec(S^*)$,  forms a feature vector of the SPD matrix, serving as a feature for classification. In summary, by leveraging the Riemannian manifold and associated mathematical tools, we can effectively analyze multivariate biosignals, compute meaningful statistics, and classification tasks commonly performed in Euclidean space with high accuracy and efficiency.

\subsubsection{Classifier}
The Radial Basis Function Support Vector Machine (RBF-SVM) is a powerful classification algorithm for non-linearly separable datasets \cite{learningwithKernels}. The RBF kernel function computes the inner-product between two vectors projected onto an infinite dimensional Euclidean space as, 
\begin{equation}
K(s, s_i) = \exp(-\gamma \cdot ||s - s_i||^2).
\end{equation}
Intuitively, it measures the similarity between data point \(s\) and support vector \(s_i\), with \(\gamma\) being the length-scale or ``spread'' of the kernel. Once can then linearly separate the projected vectors with a simple decision function defined as:
\begin{equation}
y = \text{sign}\left(\sum_{i=1}^{M} (\alpha_i \cdot y_i \cdot K(s, s_i)) + b\right)
\end{equation}
where \(y\) represents the predicted label, \(M\) is the number of support vectors, \(\alpha_i\) are Lagrange multipliers, \(y_i\) are class labels, \(K(s, s_i)\) is the RBF kernel, and \(b\) is the bias term. This equation computes a weighted sum of kernel values between input \(s\) and support vectors \(s_i\) using learned weights (\(\alpha_i\)) to make binary classifications, effectively mapping EEG signal features to labels (left or right arm movement).

\subsection{Signal Processing Pipelines}
Next we describe the process for deriving predictions, denoted as $y \in [left, right]$, from raw time-windowed EEG signals, $X \in \mathbb{R}^{W \times 30}$. We will subsequently detail the acquisition of EEG signals $X$, and the utilization of classification results $y$ for robotic arm control.

\subsubsection{EEG Feature Extraction}
We denote a sequence of raw EEG data consisting of 30 channels, which initiates at time $t$ and spans a duration of $T$, as $X_t \in \mathbb{R}^{W \times 30}$. To ensure optimal compatibility with both our classifier and robot control, we resample the data from its original sampling rate of 250 Hz to a new rate of 160 Hz for computational efficiency and to match motion capture frequency. Given the significant amplitude variability in EEG data across different subjects and trials, we introduce a data processing technique involving signal differentiation, $\hat{X}_t = X_{t} - X_{t-1}$ , to standardize the data. While less conventional, this approach aims to standardize the data by preserving essential signal tendencies related to brain activities such as readiness potential (RP) and event-related desynchronization/synchronization (ERD/ERS). 

Following the preprocessing procedures, we employ the sample covariance matrix method, discussed in Section \ref{sec:preliminaries}, to derive the feature denoted as $s_t \in \mathbb{R}^{465}$ where 465 corresponds to the flattened upper triangle of the tangent space covariance matrix denoted as $C \in \mathbb{R}^{30 \times 30}$. This particular feature is utilized consistently across all methodologies presented in this paper; thus, any reference to ``feature extraction" in the subsequent discussions pertains to it.

\subsubsection{Offline Data Collection Pipeline}
In this pipeline, we align EEG data with OptiTrack data at a frequency of 160 Hz to generate our data both for training and interpretation. As shown in Figure \ref{fig:pipeline}, our dataset consists of following parts:
\begin{equation}
\label{eq:dataset}
    DS = \{\hat{\mathbf{X}}, \mathbf{y}, \mathbf{M}\}
\end{equation}
The raw EEG signal $\mathbf{X}_{raw}$ has the shape $(T, 30)$. We apply a time window of size $W$ with step size $1$ to segment it and concatenate the windows together, resulting in an overlapped windowed data $\hat{\mathbf{X}} \in \mathbb{R}^{T \times W \times 30}$, $\mathbf{y} \in \mathbb{R}^T$ is the corresponding label  $y \in [left, right]~\forall t=1\dots T$, distinguished by the auditory cues during data collection. $\mathbf{M} \in \mathbb{R}^{T \times 2 \times 3}$ denotes motion tracking for left and right arm, translation movement of them, which is used to identify the onset of motions.
Upon establishing the dataset, we process the EEG signals $\hat{\mathbf{X}}$ through the feature extraction phase to obtain $\mathbf{s}\in\mathbb{R}^{T_{L/R} \times 465}$ where $T_{L/R}$ is the number of data points we select before or after the onset of left or right arm motion. These features $\mathbf{s}$, alongside the respective labels $\mathbf{y}$, are then employed to train the RBF-SVM classifier. 
\subsection{Online Robot-in-the-Loop Motion Prediction Pipeline}
Once the SVM classifier is trained from the offline collected data, we use the online EEG data stream to predict intention of moving left or right arm and trigger a robot to perform an assistive reaching task. At every time step, we extract time-windowed of EEG data $\hat{X}_t \in \mathbb{R}^{W \times 30}$ (with same window-size as training) and perform the same sample covariance feature extraction from training to get the corresponding feature vector $s_t \in \mathbb{R}^{465}$ for each time-step $t$. Due to non-stationarity of EEG signals and the fact that now we perform this classification with a robot-in-the-loop, the EEG signals may differ greatly compared to the offline collected data as now the human also has expectations of the motion of the robot. This can lead to erratic predictions not amenable for reactive and assistive control. To alleviate this, we introduce two novel methods designed to enhance the performance of real-time robot assistive control.

\begin{figure*}
  \begin{minipage}[b]{0.65\textwidth}
  \begin{center}
\resizebox{\linewidth}{!}{\begin{tabular}{|c|clllllllll|}
\hline
\multicolumn{1}{|l|}{} &
  \multicolumn{10}{c|}{\textbf{Frequency Range (Hz)}} \\ \hline
\multirow{8}{*}{\textbf{Time Window (s)}} &
  \multicolumn{1}{l|}{} &
  \multicolumn{1}{l|}{\textbf{0 - 5}} &
  \multicolumn{1}{l|}{\textbf{0 - 10}} &
  \multicolumn{1}{l|}{\textbf{5 - 15}} &
  \multicolumn{1}{l|}{\textbf{10 - 20}} &
  \multicolumn{1}{l|}{\textbf{15 - 25}} &
  \multicolumn{1}{l|}{\textbf{20 - 30}} &
  \multicolumn{1}{l|}{\textbf{25 - 35}} &
  \multicolumn{1}{l|}{\textbf{30 - 40}} &
  \textbf{35 - 45} \\ \cline{2-11} 
 &
  \multicolumn{1}{c|}{\textbf{0.03}} &
  \multicolumn{1}{l|}{32.47} &
  \multicolumn{1}{l|}{43.82} &
  \multicolumn{1}{l|}{\textbf{61.80}} &
  \multicolumn{1}{l|}{33.61} &
  \multicolumn{1}{l|}{16.54} &
  \multicolumn{1}{l|}{38.18} &
  \multicolumn{1}{l|}{31.86} &
  \multicolumn{1}{l|}{19.92} &
  6.75 \\ \cline{2-11} 
 &
  \multicolumn{1}{c|}{\textbf{0.06}} &
  \multicolumn{1}{l|}{44.76} &
  \multicolumn{1}{l|}{41.65} &
  \multicolumn{1}{l|}{\textbf{58.40}} &
  \multicolumn{1}{l|}{26.56} &
  \multicolumn{1}{l|}{15.34} &
  \multicolumn{1}{l|}{28.76} &
  \multicolumn{1}{l|}{18.36} &
  \multicolumn{1}{l|}{15.92} &
  37.41 \\ \cline{2-11} 
 &
  \multicolumn{1}{c|}{\textbf{0.12}} &
  \multicolumn{1}{l|}{12.24} &
  \multicolumn{1}{l|}{29.53} &
  \multicolumn{1}{l|}{\textbf{69.80}} &
  \multicolumn{1}{l|}{43.34} &
  \multicolumn{1}{l|}{22.60} &
  \multicolumn{1}{l|}{10.47} &
  \multicolumn{1}{l|}{4.44} &
  \multicolumn{1}{l|}{9.62} &
  19.34 \\ \cline{2-11} 
 &
  \multicolumn{1}{c|}{\textbf{0.25}} &
  \multicolumn{1}{l|}{42.99} &
  \multicolumn{1}{l|}{45.92} &
  \multicolumn{1}{l|}{\textbf{63.82}} &
  \multicolumn{1}{l|}{32.91} &
  \multicolumn{1}{l|}{6.25} &
  \multicolumn{1}{l|}{9.40} &
  \multicolumn{1}{l|}{15.11} &
  \multicolumn{1}{l|}{34.25} &
  12.57 \\ \cline{2-11} 
 &
  \multicolumn{1}{c|}{\textbf{0.5}} &
  \multicolumn{1}{l|}{15.49} &
  \multicolumn{1}{l|}{33.22} &
  \multicolumn{1}{l|}{58.705} &
  \multicolumn{1}{l|}{7.27} &
  \multicolumn{1}{l|}{8.63} &
  \multicolumn{1}{l|}{27.42} &
  \multicolumn{1}{l|}{34.79} &
  \multicolumn{1}{l|}{\textbf{61.51}} &
  6.51 \\ \cline{2-11} 
 &
  \multicolumn{1}{c|}{\textbf{1}} &
  \multicolumn{1}{l|}{34.01} &
  \multicolumn{1}{l|}{39.33} &
  \multicolumn{1}{l|}{57.91} &
  \multicolumn{1}{l|}{10.37} &
  \multicolumn{1}{l|}{28.69} &
  \multicolumn{1}{l|}{48.15} &
  \multicolumn{1}{l|}{60.17} &
  \multicolumn{1}{l|}{\textbf{80.11}} &
  71.27 \\ \cline{2-11} 
 &
  \multicolumn{1}{c|}{\textbf{2}} &
  \multicolumn{1}{l|}{11.36} &
  \multicolumn{1}{l|}{39.03} &
  \multicolumn{1}{l|}{\cellcolor{blue!15} \textbf{86.88}} &
  \multicolumn{1}{l|}{\cellcolor{blue!15}  85.13} &
  \multicolumn{1}{l|}{79.19} &
  \multicolumn{1}{l|}{73.02} &
  \multicolumn{1}{l|}{60.29} &
  \multicolumn{1}{l|}{42.51} &
  39.43 \\ \hline
\end{tabular}}
\captionof{table}{Mean accuracy of SVM classification on testing datasets across various frequency ranges (columns) and time windows (rows).}
\label{tab:hyperparam_selection}
\end{center}
\end{minipage}
\begin{minipage}[b]{0.325\textwidth}
\resizebox{\linewidth}{!}{
\begin{tabular}{|c|ccc|}
\hline
 & \multicolumn{1}{c|}{\textbf{Raw}} & \multicolumn{1}{c|}{\textbf{Covariance}} & \textbf{Tangent space} \\ \hline
Accuracy (\%) w/o derivative & \multicolumn{1}{c|}{32.84} & \multicolumn{1}{c|}{47.27} & 40.32  \\ \hline
\textbf{Accuracy (\%) with derivative} & \multicolumn{1}{c|}{64.15} & \multicolumn{1}{c|}{56.26} & \cellcolor{blue!15} \textbf{86.88}\\ \hline
\end{tabular}}
\vspace{-5pt}
\captionof{table}{SVM performance across features}\label{tab:feature-selection}
\centering

\resizebox{0.7\linewidth}{!}{\begin{tabular}{|c|c|}
\hline
 \textbf{Classifier} & Accuracy (\%) \\ \hline
Linear Regressor & 63.12 \\ \hline
\textbf{Support Vector Machine} & \cellcolor{blue!15} \textbf{69.10} \\ \hline
Multi-Layer Perceptron & 45.93 \\ \hline
Random Forest & 47.47 \\ \hline
\end{tabular}}
\captionof{table}{Average of features accuracy (from Table \ref{tab:feature-selection}) with different classifiers}\label{tab:model-selection}
\end{minipage}
\vspace{-15pt}
\end{figure*}

\textbf{Score Thresholding}: 
During the prediction phase, our approach generates classification scores, $P_t \in \mathbb{R}^2$, representing the probability of the `left' and `right' classes. These scores are constrained such that they sum to 1. However, rather than directly making predictions based on these scores, we apply a decision threshold $\delta$. This strategy ensures that predictions are made only when confident; i.e., when the probability of a class exceeds the threshold, thereby mitigating the impact of noise present in EEG data.

\textbf{Buffer Queuing}: Despite the incorporation of score thresholding, occasional classification errors can affect the precision of robot control. For instance, in scenarios where both correct and incorrect predictions co-exist, it is imperative to address the potential dominance of sporadic errors.
To refine control precision, we implement a buffer queue, $\mathcal{BQ}$, which retains the latest $q$ classification outcomes. This buffer is recurrently updated, and when a motion command is imminent, the most frequently occurring prediction within $\mathcal{BQ}$ is chosen as the directive $x^{\star}$ for robot actuation.

\textbf{Robot Control} Once `left' or `right' motor intention is predicted with confidence the target of the end-effector of the robot $x^*\in\mathbb{R}^3$ is set to a pre-defined location of objects on the left and right side of the human. Once the target for the robot’s end-effector is defined, we trigger a linear controller of the following form $\dot{x} = A(x-x^*)$ with $A\prec 0$ and $x,\dot{x}\in\mathbb{R}^3$ being the current end-effector position and desired velocity. This simple controller will drive the robot to the desired target once human motor intention is predicted.

\section{Classifier Evaluation and Analysis}
\label{sec:experiments}
We begin by describing the dataset used in our experiments in Section \ref{sec:dataset}. We then present a hyper-parameter evaluation of our proposed Riemannian based EEG feature in search of the optimal the window size $W$ and frequency range, elucidating the intrinsic relationship between different brain frequency bands and motor movements in Section \ref{sec:hyper}. Based on the optimal $W$, we compare EEG features with or without the derivative method, including the raw signal, sample covariance matrix, and projection to tangent space in Section \ref{sec:feature}. We also evaluate classification accuracy of different classifiers, including logistic regression, random forest, multi-layer perceptron, and SVM. Results of these assessments can be found in Table \ref{tab:hyperparam_selection}, \ref{tab:feature-selection} and \ref{tab:model-selection}, concluding that a time window of 2s with EEG derivatives and sample covariance matrix projected on the tangent space using SVM for classification yields the best performance.
\vspace{-2.5pt}
\subsection{Description of self-collected datasets}
\label{sec:dataset}
We use the self-collected offline dataset to select the best combination of classfiers, features and hyperparameters. This offline dataset is different from the dataset used for real robot experiment. We use a Bittium Neurone \cite{Neurone} wet EEG device to get a stream of brain signals at 250 Hz, which is downsampled to 160 Hz, using the International 10–20 system with 30 electrodes and Optitrack motion tracking system at 100 Hz to collect data offline and conduct real robot experiments online. Two subjects perform tasks involving grasping objects with their left and right hands. Each subject repeats each task 30 times during two recordings to gather the training dataset, guided by auditory cues. Each task consists of a 5-second rest period followed by a 5-second action period. The same process is followed to collect the testing dataset, which takes approximately 10 minutes for each training and testing dataset. Although we provide auditory cues to guide the subjects, we also use Optitrack body tracking to precisely determine when each motion starts and stops. This helps us slice the data accurately between 1 seconds before and until the end of motion. Any data before this one-second mark is considered the resting state.
\vspace{-5pt}
\subsection{Time-Window Hyper-Parameters Selection}
\label{sec:hyper}
We evaluate the classification accuracy of our proposed time-window based feature classifier, with a particular focus on two hyper-parameters: time window and frequency range. Our analysis employs the RBF SVM model, and the outcomes are delineated in Table \ref{tab:hyperparam_selection}. Our findings reveal that utilizing a frequency range of 5-15 Hz yielded the most favorable results, which we presume to be attributed by the presence of RP occurring within the 8-13 Hz range \cite{pineda2000effects}. We observed promising outcomes upon extending the time window to 0.5 and 1 second within the 30-40 Hz gamma range, which aligns with its known association with advanced cognitive functions. Nonetheless, across our comprehensive analysis, the 5-15 Hz frequency range consistently demonstrates superior performance with an optimal choice of a 2-second time window, resulting in an impressive accuracy of 86.88\%. This highlights the importance of fine-tuning hyperparameters to optimize EEG signal analysis.
\vspace{-5pt}
\subsection{Feature and Classifier Selection}
\label{sec:feature}
Next, we provide a comparative analysis of classification accuracy using different classifiers with diverse feature sets. These feature sets encompass the raw signal data of windowed 30-electrode signals, the upper triangle of the covariance matrix, and the tangent space projection, all with and without derivative processing. Results are presented in Table \ref{tab:feature-selection} (feature selection) and Table \ref{tab:model-selection} (classifier selection). Upon evaluating the feature selection outcomes, it becomes evident that incorporating the derivative processing significantly enhances overall accuracy. This enhancement stems from its capacity to mitigate inter-trial disparities, resulting in a notable increase of 28.96 percentage points and yielding an accuracy of 69.10\%. When considering feature types, the tangent space projection method emerges as the most effective, achieving the highest accuracy at 86.88\%. Regarding the selection of classifiers, as depicted in Table \ref{tab:model-selection}, we compute the average of the top 3 accuracy from all 6 feature combinations (as shown in Table \ref{tab:feature-selection}) for each classifier to gauge their overall performance. Optimal hyper-parameters for SVM ($C=0.1,\gamma=0.5$) and MLP (hidden layer size $(100, 3)$) were obtained via grid search. As shown, SVM outperforms, attaining an average accuracy of 69.1\%.

\section{Robot-in-the-Loop Validation and Analysis}
\label{sec:robot_exp_biological}
\subsection{Real robot testing and results}
For real robot testing, we follow a similar offline data collection stage as in the evaluation phase with a KUKA iiwa 7 robot. We gather 30 randomized trials of left and right grasping for two subjects. However, each trial now consists of a 2s rest period followed by a 2s motion period. Once the data is collected from the offline pipeline, we employ it to train the SVM with the predetermined feature, tangent space projection of sample covariance matrix, and hyper-parameters, 2s window and 5-15 Hz. We the validate the trained model online, where our experiment setup assumes a robot and a human mirroring each other in a shared workspace with two objects in front of the human, as in Fig. \ref{fig:pipeline}. In response to an auditory cue (``move") signaling motor initiation, subjects execute randomized grasping motions using either the left or right hand. The robot, based solely on the prediction of the EEG signal given by the trained SVM, moves in that direction and gives the object in that direction to the human, which is not reachable by the human.

\begin{figure}[!tbp]
  \centering
  \includegraphics[width=\linewidth]{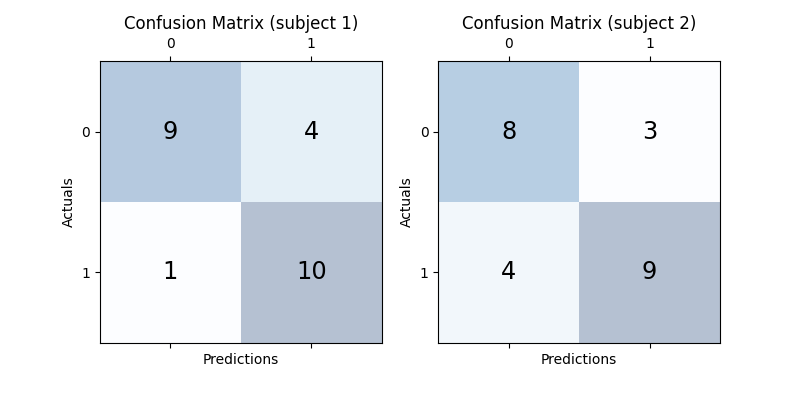}
  \vspace{-20pt}
  \caption{Confusion matrix of robot-in-the-loop experiments \label{fig:confusion_matrix_robot}}
    \vspace{-15pt}
\end{figure}

The accuracy of the two different subjects differed (subject 1: 79.1\%, subject 2: 70.8\%) but was above 70\% as shown in Fig. \ref{fig:confusion_matrix_robot} for 24 trials with random movement from the subject. Since the classification result of the left and right movements is prone to noise due to the resting state and readiness potential timing error prior to the onset of motion, we used a queue, which increased the accuracy of the classification. We further evaluated the contribution of different brain regions by separating the signals into frontal lobe (FCz, FC3, FC4, FT7, FT8), central lobe (Cz, C3, C4, T7, T8), parietal lobe (CPz, CP3, CP4, TP7, TP8), and occipital lobe (Pz, P3, P4, P7, P8), then train the SVM separately. The results show the highest accuracy on the parietal lobe (72.77\%), followed by the central lobe (60.53\%), as indicated in the literature \cite{Wang2005CommonSP}.

\begin{figure}[!tbp]
  \centering
  \includegraphics[trim={0 0 0cm 1cm},clip,width=0.75\linewidth]{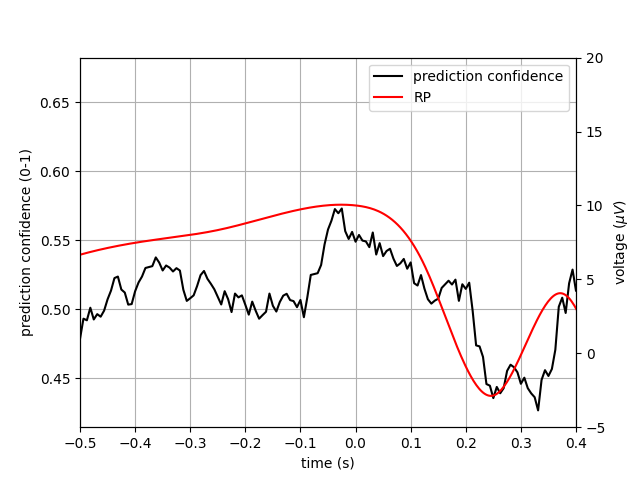}
  \caption{Mean probability confidence for all left and right trials in black line with respect to the onset of motion and the readiness potential (RP) in red line in $\mu V$ units reproduced from \cite{Wen2018TheRP}.}
  \vspace{-15pt}
  \label{readiness_potential}
\end{figure}

\subsection{Biological Interpretation}By aggregating the mean probability (classification score) generated by the SVM for left and right motions across each trial and aligning it with the motion timing data obtained from the motion capture system, we construct the black curve in Fig. \ref{readiness_potential}. The x-axis at `0s' represents the onset of motion. We observe a notable increase in signal intensity from approximately -0.1s to 0s, followed by a subsequent decrease from 0s to 0.3s. These observations potentially suggest the presence of the readiness potential (RP), the red curve. However, while we detect RP-like behavior, it’s difficult to disambiguate the onset of motion in a real-time setting given the current technology. This is due to two primary reasons: the RP signal is very brief, and EEG also detects motion artifacts. We believe that we are detecting a combination of RP and motion-related brain signals at 5-15 Hz; still useful for control as it has both preparatory brain activity and the intention to execute a specific motion.

\section{Conclusion and Future Work}
\label{sec:conclusion}
In this study, we present two distinct pipelines tailored to facilitate EEG-based real-time robot control. The online pipeline  primarily focuses on feature extraction from raw EEG data for classification purposes. To address the inherent challenge of noisy signals, we introduce classification score thresholding and buffer queuing methods, which enhance signal stability, enabling control of the robot arm in real-time. The offline pipeline is responsible for the collection and processing of EEG data used for training our classifier. Additionally, this pipeline handles the acquisition of motion data, which is later analyzed for biological interpretation.

Our experimental results demonstrate the effectiveness of the tangent space covariance matrix projection method in extracting robust features that are compatible with multiple classifiers. Among these classifiers, the support vector machine (SVM) exhibits the highest performance. Furthermore, we conduct a comprehensive evaluation of various parameters, including time window size, frequency bands, EEG data features, and classifiers. This analysis allows us to identify the most effective combination for online classification. Leveraging the motion capture system, we investigate the temporal relationship between motion and classification, shedding light on the influence of readiness potential and motor brain signals on the classification outcomes.

In the future, our research direction involves the integration of additional modalities, such as electromyography (EMG), with the aim of improving the quality of neural data. This enhancement may enable us to utilize stable intention signals for robot control even before the commencement of physical movements. Furthermore, we aspire to expand our classification repertoire beyond left and right grasping data, incorporating more complex movements involving the arms or legs, thereby broadening the scope of assistive control.






\clearpage
\bibliographystyle{IEEEtran}
\bibliography{citations.bib}

\begin{thebibliography}{10}
\providecommand{\url}[1]{#1}
\csname url@samestyle\endcsname
\providecommand{\newblock}{\relax}
\providecommand{\bibinfo}[2]{#2}
\providecommand{\BIBentrySTDinterwordspacing}{\spaceskip=0pt\relax}
\providecommand{\BIBentryALTinterwordstretchfactor}{4}
\providecommand{\BIBentryALTinterwordspacing}{\spaceskip=\fontdimen2\font plus
\BIBentryALTinterwordstretchfactor\fontdimen3\font minus
  \fontdimen4\font\relax}
\providecommand{\BIBforeignlanguage}[2]{{%
\expandafter\ifx\csname l@#1\endcsname\relax
\typeout{** WARNING: IEEEtran.bst: No hyphenation pattern has been}%
\typeout{** loaded for the language `#1'. Using the pattern for}%
\typeout{** the default language instead.}%
\else
\language=\csname l@#1\endcsname
\fi
#2}}
\providecommand{\BIBdecl}{\relax}
\BIBdecl

\bibitem{Sciutti2018HumanizingHI}
\BIBentryALTinterwordspacing
A.~Sciutti, M.~Mara, V.~Tagliasco, and G.~Sandini, ``Humanizing human-robot
  interaction: On the importance of mutual understanding,'' \emph{IEEE
  Technology and Society Magazine}, vol.~37, pp. 22--29, 2018. [Online].
  Available: \url{https://api.semanticscholar.org/CorpusID:3812272}
\BIBentrySTDinterwordspacing

\bibitem{Kuli2003EstimatingIF}
\BIBentryALTinterwordspacing
D.~Kuli{\'c} and E.~A. Croft, ``Estimating intent for human-robot
  interaction,'' \emph{Advanced Robotics}, 2003. [Online]. Available:
  \url{https://api.semanticscholar.org/CorpusID:1120056}
\BIBentrySTDinterwordspacing

\bibitem{ATHAVALE201722}
\BIBentryALTinterwordspacing
Y.~Athavale and S.~Krishnan, ``Biosignal monitoring using wearables:
  Observations and opportunities,'' \emph{Biomedical Signal Processing and
  Control}, vol.~38, pp. 22--33, 2017. [Online]. Available:
  \url{https://www.sciencedirect.com/science/article/pii/S1746809417300617}
\BIBentrySTDinterwordspacing

\bibitem{Esposito2021-to}
D.~Esposito, J.~Centracchio, E.~Andreozzi, G.~D. Gargiulo, G.~R. Naik, and
  P.~Bifulco, ``\BIBforeignlanguage{en}{{Biosignal-Based} {Human-Machine}
  interfaces for assistance and rehabilitation: A survey},''
  \emph{\BIBforeignlanguage{en}{Sensors (Basel)}}, vol.~21, no.~20, Oct. 2021.

\bibitem{TANG2023108712}
\BIBentryALTinterwordspacing
C.~Tang, Z.~Xu, E.~Occhipinti, W.~Yi, M.~Xu, S.~Kumar, G.~S. Virk, S.~Gao, and
  L.~G. Occhipinti, ``From brain to movement: Wearables-based motion intention
  prediction across the human nervous system,'' \emph{Nano Energy}, vol. 115,
  p. 108712, 2023. [Online]. Available:
  \url{https://www.sciencedirect.com/science/article/pii/S2211285523005499}
\BIBentrySTDinterwordspacing

\bibitem{mohebbi_human-robot_2020}
\BIBentryALTinterwordspacing
A.~Mohebbi, ``Human-{Robot} {Interaction} in {Rehabilitation} and {Assistance}:
  a {Review},'' \emph{Current Robotics Reports}, vol.~1, no.~3, pp. 131--144,
  Sep. 2020. [Online]. Available:
  \url{https://doi.org/10.1007/s43154-020-00015-4}
\BIBentrySTDinterwordspacing

\bibitem{Lakany2007UnderstandingIO}
\BIBentryALTinterwordspacing
H.~M. Lakany and B.~A. Conway, ``Understanding intention of movement from
  electroencephalograms,'' \emph{Expert Systems}, vol.~24, 2007. [Online].
  Available: \url{https://api.semanticscholar.org/CorpusID:5936372}
\BIBentrySTDinterwordspacing

\bibitem{kirschstein2009source}
T.~Kirschstein and R.~K{\"o}hling, ``What is the source of the eeg?''
  \emph{Clinical EEG and neuroscience}, vol.~40, no.~3, pp. 146--149, 2009.

\bibitem{Vrbu2022PastPA}
\BIBentryALTinterwordspacing
K.~V{\"a}rbu, M.~Naveed, and Y.~Muhammad, ``Past, present, and future of
  eeg-based bci applications,'' \emph{Sensors (Basel, Switzerland)}, vol.~22,
  2022. [Online]. Available:
  \url{https://api.semanticscholar.org/CorpusID:248775078}
\BIBentrySTDinterwordspacing

\bibitem{stages}
M.~B. Khalid, N.~I. Rao, I.~Rizwan-i Haque, S.~Munir, and F.~Tahir, ``Towards a
  brain computer interface using wavelet transform with averaged and time
  segmented adapted wavelets,'' in \emph{2009 2nd International Conference on
  Computer, Control and Communication}, 2009, pp. 1--4.

\bibitem{meng_noninvasive_2016}
\BIBentryALTinterwordspacing
J.~Meng, S.~Zhang, A.~Bekyo, J.~Olsoe, B.~Baxter, and B.~He, ``Noninvasive
  {Electroencephalogram} {Based} {Control} of a {Robotic} {Arm} for {Reach} and
  {Grasp} {Tasks},'' \emph{Scientific Reports}, vol.~6, no.~1, p. 38565, Dec.
  2016. [Online]. Available: \url{https://doi.org/10.1038/srep38565}
\BIBentrySTDinterwordspacing

\bibitem{lee2023noir}
R.~Zhang, S.~Lee, M.~Hwang, A.~Hiranaka, C.~Wang, W.~Ai, J.~J.~R. Tan,
  S.~Gupta, Y.~Hao, G.~Levine, R.~Gao, A.~Norcia, L.~Fei-Fei, and J.~Wu,
  ``Noir: Neural signal operated intelligent robots for everyday activities,''
  in \emph{7th Annual Conference on Robot Learning}, 2023.

\bibitem{Dillen2022-ti}
A.~Dillen, D.~Steckelmacher, K.~Efthymiadis, K.~Langlois, A.~De~Beir,
  U.~Marusic, B.~Vanderborght, A.~Now{\'e}, R.~Meeusen, F.~Ghaffari, O.~Romain,
  and K.~De~Pauw, ``\BIBforeignlanguage{en}{Deep learning for biosignal
  control: insights from basic to real-time methods with recommendations},''
  \emph{\BIBforeignlanguage{en}{J Neural Eng}}, vol.~19, no.~1, Feb. 2022.

\bibitem{100}
F.~S. Racz, R.~Fakhreddine, S.~Kumar, and J.~Del R.~Millan, ``Riemannian
  geometry-based detection of slow cortical potentials during movement
  preparation,'' in \emph{2023 11th International IEEE/EMBS Conference on
  Neural Engineering (NER)}, 2023, pp. 1--5.

\bibitem{7740054}
F.~Yger, M.~Berar, and F.~Lotte, ``Riemannian approaches in brain-computer
  interfaces: A review,'' \emph{IEEE Transactions on Neural Systems and
  Rehabilitation Engineering}, vol.~25, no.~10, pp. 1753--1762, 2017.

\bibitem{ying_riemannian_2022}
\BIBentryALTinterwordspacing
J.~Ying, Q.~Wei, and X.~Zhou, ``Riemannian geometry-based transfer learning for
  reducing training time in c-{VEP} {BCIs},'' \emph{Scientific Reports},
  vol.~12, no.~1, p. 9818, Jun. 2022. [Online]. Available:
  \url{https://doi.org/10.1038/s41598-022-14026-y}
\BIBentrySTDinterwordspacing

\bibitem{Congedo2017RiemannianGF}
\BIBentryALTinterwordspacing
M.~Congedo, A.~Barachant, and R.~Bhatia, ``Riemannian geometry for eeg-based
  brain-computer interfaces; a primer and a review,'' 2017. [Online].
  Available: \url{https://api.semanticscholar.org/CorpusID:13857467}
\BIBentrySTDinterwordspacing

\bibitem{Rodrigues2019RiemannianPA}
\BIBentryALTinterwordspacing
P.~L.~C. Rodrigues, C.~Jutten, and M.~Congedo, ``Riemannian procrustes
  analysis: Transfer learning for brain–computer interfaces,'' \emph{IEEE
  Transactions on Biomedical Engineering}, vol.~66, pp. 2390--2401, 2019.
  [Online]. Available: \url{https://api.semanticscholar.org/CorpusID:58642888}
\BIBentrySTDinterwordspacing

\bibitem{Barachant2012MulticlassBI}
\BIBentryALTinterwordspacing
A.~Barachant, S.~Bonnet, M.~Congedo, and C.~Jutten, ``Multiclass
  brain–computer interface classification by riemannian geometry,''
  \emph{IEEE Transactions on Biomedical Engineering}, vol.~59, pp. 920--928,
  2012. [Online]. Available:
  \url{https://api.semanticscholar.org/CorpusID:423006}
\BIBentrySTDinterwordspacing

\bibitem{Barachant2013ClassificationOC}
\BIBentryALTinterwordspacing
------, ``Classification of covariance matrices using a riemannian-based kernel
  for bci applications,'' \emph{Neurocomputing}, vol. 112, pp. 172--178, 2013.
  [Online]. Available: \url{https://api.semanticscholar.org/CorpusID:13873072}
\BIBentrySTDinterwordspacing

\bibitem{Michel2019EEGSI}
\BIBentryALTinterwordspacing
C.~M. Michel and D.~Brunet, ``Eeg source imaging: A practical review of the
  analysis steps,'' \emph{Frontiers in Neurology}, vol.~10, 2019. [Online].
  Available: \url{https://api.semanticscholar.org/CorpusID:93003798}
\BIBentrySTDinterwordspacing

\bibitem{RP/ERD}
Y.~Wang, S.~Gao, and X.~Gao, ``Common spatial pattern method for channel
  selelction in motor imagery based brain-computer interface,'' in \emph{2005
  IEEE Engineering in Medicine and Biology 27th Annual Conference}, 2005, pp.
  5392--5395.

\bibitem{Andreou2016ComparingEI}
\BIBentryALTinterwordspacing
D.~A. Andreou and R.~Poli, ``Comparing eeg, its time-derivative and their joint
  use as features in a bci for 2-d pointer control,'' \emph{2016 38th Annual
  International Conference of the IEEE Engineering in Medicine and Biology
  Society (EMBC)}, pp. 5853--5856, 2016. [Online]. Available:
  \url{https://api.semanticscholar.org/CorpusID:172677}
\BIBentrySTDinterwordspacing

\bibitem{Menceloglu2020SpectralpowerAR}
\BIBentryALTinterwordspacing
M.~Menceloglu, M.~Grabowecky, and S.~Suzuki, ``Spectral-power associations
  reflect amplitude modulation and within-frequency interactions on the
  sub-second timescale and cross-frequency interactions on the seconds
  timescale,'' \emph{PLoS ONE}, vol.~15, 2020. [Online]. Available:
  \url{https://api.semanticscholar.org/CorpusID:214419358}
\BIBentrySTDinterwordspacing

\bibitem{Chen2022MentalSC}
\BIBentryALTinterwordspacing
G.~Chen, H.~S. Helm, K.~Lytvynets, W.~Yang, and C.~E. Priebe, ``Mental state
  classification using multi-graph features,'' \emph{Frontiers in Human
  Neuroscience}, vol.~16, 2022. [Online]. Available:
  \url{https://api.semanticscholar.org/CorpusID:247187688}
\BIBentrySTDinterwordspacing

\bibitem{BI2019113}
\BIBentryALTinterwordspacing
L.~Bi, A.~G. Feleke, and C.~Guan, ``A review on emg-based motor intention
  prediction of continuous human upper limb motion for human-robot
  collaboration,'' \emph{Biomedical Signal Processing and Control}, vol.~51,
  pp. 113--127, 2019. [Online]. Available:
  \url{https://www.sciencedirect.com/science/article/pii/S1746809419300473}
\BIBentrySTDinterwordspacing

\bibitem{trigili2019detection}
E.~Trigili, L.~Grazi, S.~Crea, A.~Accogli, J.~Carpaneto, S.~Micera,
  N.~Vitiello, and A.~Panarese, ``Detection of movement onset using emg signals
  for upper-limb exoskeletons in reaching tasks,'' \emph{Journal of
  neuroengineering and rehabilitation}, vol.~16, pp. 1--16, 2019.

\bibitem{SCHURGER2021558}
\BIBentryALTinterwordspacing
A.~Schurger, P.~B. Hu, J.~Pak, and A.~L. Roskies, ``What is the readiness
  potential?'' \emph{Trends in Cognitive Sciences}, vol.~25, no.~7, pp.
  558--570, 2021. [Online]. Available:
  \url{https://www.sciencedirect.com/science/article/pii/S1364661321000930}
\BIBentrySTDinterwordspacing

\bibitem{TRAVERS2020116286}
\BIBentryALTinterwordspacing
E.~Travers, N.~Khalighinejad, A.~Schurger, and P.~Haggard, ``Do readiness
  potentials happen all the time?'' \emph{NeuroImage}, vol. 206, p. 116286,
  2020. [Online]. Available:
  \url{https://www.sciencedirect.com/science/article/pii/S1053811919308778}
\BIBentrySTDinterwordspacing

\bibitem{Schurger2021WhatIT}
\BIBentryALTinterwordspacing
A.~Schurger, P.~Hu, J.~Pak, and A.~L. Roskies, ``What is the readiness
  potential?'' \emph{Trends in cognitive sciences}, vol.~25, pp. 558 -- 570,
  2021. [Online]. Available:
  \url{https://api.semanticscholar.org/CorpusID:233471654}
\BIBentrySTDinterwordspacing

\bibitem{114}
X.~Tang, W.~Li, X.~Li, W.~Ma, and X.~Dang, ``Motor imagery eeg recognition
  based on conditional optimization empirical mode decomposition and
  multi-scale convolutional neural network,'' \emph{Expert Systems with
  Applications}, vol. 149, p. 113285, 2020.

\bibitem{115}
P.~Batres-Mendoza, E.~I. Guerra-Hernandez, A.~Espinal, E.~P{\'e}rez-Careta, and
  H.~Rostro-Gonzalez, ``Biologically-inspired legged robot locomotion
  controlled with a bci by means of cognitive monitoring,'' \emph{IEEE Access},
  vol.~9, pp. 35\,766--35\,777, 2021.

\bibitem{101}
S.~Bhattacharyya, A.~Konar, and D.~Tibarewala, ``Motor imagery and error
  related potential induced position control of a robotic arm,'' \emph{IEEE/CAA
  Journal of Automatica Sinica}, vol.~4, no.~4, pp. 639--650, 2017.

\bibitem{102}
J.~Choi, K.~T. Kim, J.~H. Jeong, L.~Kim, S.~J. Lee, and H.~Kim, ``Developing a
  motor imagery-based real-time asynchronous hybrid bci controller for a
  lower-limb exoskeleton,'' \emph{Sensors}, vol.~20, no.~24, p. 7309, 2020.

\bibitem{103}
H.~Gao, L.~Luo, M.~Pi, Z.~Li, Q.~Li, K.~Zhao, and J.~Huang, ``Eeg-based
  volitional control of prosthetic legs for walking in different terrains,''
  \emph{IEEE Transactions on Automation Science and Engineering}, vol.~18,
  no.~2, pp. 530--540, 2019.

\bibitem{105}
J.~Andreu-Perez, F.~Cao, H.~Hagras, and G.-Z. Yang, ``A self-adaptive online
  brain--machine interface of a humanoid robot through a general type-2 fuzzy
  inference system,'' \emph{IEEE Transactions on Fuzzy Systems}, vol.~26,
  no.~1, pp. 101--116, 2016.

\bibitem{106}
L.~Tonin, F.~C. Bauer, and J.~d.~R. Mill{\'a}n, ``The role of the control
  framework for continuous teleoperation of a brain--machine interface-driven
  mobile robot,'' \emph{IEEE Transactions on Robotics}, vol.~36, no.~1, pp.
  78--91, 2019.

\bibitem{107}
D.~Liu, W.~Chen, Z.~Pei, and J.~Wang, ``A brain-controlled lower-limb
  exoskeleton for human gait training,'' \emph{Review of Scientific
  Instruments}, vol.~88, no.~10, 2017.

\bibitem{109}
J.~R. Millan and J.~Mouri{\~n}o, ``Asynchronous bci and local neural
  classifiers: an overview of the adaptive brain interface project,''
  \emph{IEEE transactions on neural systems and rehabilitation engineering},
  vol.~11, no.~2, pp. 159--161, 2003.

\bibitem{110}
L.~Junwei, S.~Ramkumar, G.~Emayavaramban, M.~Thilagaraj, V.~Muneeswaran, M.~P.
  Rajasekaran, V.~Venkataraman, A.~F. Hussein \emph{et~al.}, ``Brain computer
  interface for neurodegenerative person using electroencephalogram,''
  \emph{IEEE Access}, vol.~7, pp. 2439--2452, 2018.

\bibitem{111}
T.~Li, J.~Hong, J.~Zhang, and F.~Guo, ``Brain--machine interface control of a
  manipulator using small-world neural network and shared control strategy,''
  \emph{Journal of neuroscience methods}, vol. 224, pp. 26--38, 2014.

\bibitem{112}
Z.~Tang, S.~Sun, S.~Zhang, Y.~Chen, C.~Li, and S.~Chen, ``A brain-machine
  interface based on erd/ers for an upper-limb exoskeleton control,''
  \emph{Sensors}, vol.~16, no.~12, p. 2050, 2016.

\bibitem{113}
G.~Kucukyildiz, H.~Ocak, S.~Karakaya, and O.~Sayli, ``Design and implementation
  of a multi sensor based brain computer interface for a robotic wheelchair,''
  \emph{Journal of Intelligent \& Robotic Systems}, vol.~87, pp. 247--263,
  2017.

\bibitem{Nguyen2018EEGFD}
\BIBentryALTinterwordspacing
C.~H. Nguyen and P.~K. Artemiadis, ``Eeg feature descriptors and discriminant
  analysis under riemannian manifold perspective,'' \emph{Neurocomputing}, vol.
  275, pp. 1871--1883, 2018. [Online]. Available:
  \url{https://api.semanticscholar.org/CorpusID:894293}
\BIBentrySTDinterwordspacing

\bibitem{Kalaganis2020ARG}
\BIBentryALTinterwordspacing
F.~P. Kalaganis, N.~A. Laskaris, E.~Chatzilari, S.~Nikolopoulos, and
  I.~Kompatsiaris, ``A riemannian geometry approach to reduced and
  discriminative covariance estimation in brain computer interfaces,''
  \emph{IEEE Transactions on Biomedical Engineering}, vol.~67, pp. 245--255,
  2020. [Online]. Available:
  \url{https://api.semanticscholar.org/CorpusID:122544415}
\BIBentrySTDinterwordspacing

\bibitem{Calinon2019GaussiansOR}
\BIBentryALTinterwordspacing
S.~Calinon, ``Gaussians on riemannian manifolds: Applications for robot
  learning and adaptive control,'' \emph{IEEE Robotics \& Automation Magazine},
  vol.~27, pp. 33--45, 2019. [Online]. Available:
  \url{https://api.semanticscholar.org/CorpusID:216322247}
\BIBentrySTDinterwordspacing

\bibitem{learningwithKernels}
B.~Sch\"olkopf, A.~J. Smola, and F.~Bach, \emph{Learning with Kernels: Support
  Vector Machines, Regularization, Optimization, and Beyond}.\hskip 1em plus
  0.5em minus 0.4em\relax The MIT Press, 2018.

\bibitem{Neurone}
\BIBentryALTinterwordspacing
{EEG System for Clinical and Research Use | Bittium NeurOne}. Bittium. Accessed
  2023-03-06. [Online]. Available:
  \url{https://www.bittium.com/medical/bittium-neuron}
\BIBentrySTDinterwordspacing

\bibitem{pineda2000effects}
J.~A. Pineda, B.~Z. Allison, and A.~Vankov, ``The effects of self-movement,
  observation, and imagination on/spl mu/rhythms and readiness potentials
  (rp's): toward a brain-computer interface (bci),'' \emph{IEEE Transactions on
  Rehabilitation Engineering}, vol.~8, no.~2, pp. 219--222, 2000.

\bibitem{Wang2005CommonSP}
\BIBentryALTinterwordspacing
Y.~Wang, S.~Gao, and X.~Gao, ``Common spatial pattern method for channel
  selelction in motor imagery based brain-computer interface,'' \emph{2005 IEEE
  Engineering in Medicine and Biology 27th Annual Conference}, pp. 5392--5395,
  2005. [Online]. Available:
  \url{https://api.semanticscholar.org/CorpusID:20814659}
\BIBentrySTDinterwordspacing

\bibitem{Wen2018TheRP}
\BIBentryALTinterwordspacing
W.~Wen, R.~Minohara, S.~Hamasaki, T.~Maeda, Q.~An, Y.~Tamura, H.~Yamakawa,
  A.~Yamashita, and H.~Asama, ``The readiness potential reflects the
  reliability of action consequence,'' \emph{Scientific Reports}, vol.~8, 2018.
  [Online]. Available: \url{https://api.semanticscholar.org/CorpusID:51940807}
\BIBentrySTDinterwordspacing

\end{thebibliography}

\end{document}